\documentclass[preprint,review,12pt]{elsarticle}

\usepackage{amssymb}
\usepackage{amsmath}

\usepackage{multirow}
\usepackage{url}

\journal{Pattern Recognition}

\begin{document}

\begin{frontmatter}

\title{Budget-Aware Pruning: Handling Multiple Domains with Less Parameters}

\author[1]{Samuel Felipe dos Santos\corref{cor1}} 
\cortext[cor1]{Corresponding author.
}
\ead{sam.fe.san@gmail.com}
\author[2]{Rodrigo Berriel}
\ead{berriel@lcad.inf.ufes.br}
\author[2]{Thiago Oliveira-Santos}
\ead{todsantos@inf.ufes.br}
\author[3]{Nicu Sebe}
\ead{niculae.sebe@unitn.it}
\author[1]{Jurandy Almeida}
\ead{jurandy.almeida@ufscar.br}

\affiliation[1]{
                organization={Federal University of São Carlos (UFSCar)},
                city={São Carlos}, 
                country={Brazil}}

\affiliation[2]{
                organization={Federal University of Espírito Santo (UFES)},
                city={Vitória}, 
                country={Brazil}}

\affiliation[3]{
                organization={University of Trento (UniTN)},
                city={Trento},
                country={Italy}}

\begin{abstract}
Deep learning has achieved state-of-the-art performance on several computer vision tasks and domains. Nevertheless, it still has a high computational cost and demands a significant amount of parameters. Such requirements hinder the use in resource-limited environments and demand both software and hardware optimization. Another limitation is that deep models are usually specialized into a single domain or task, requiring them to learn and store new parameters for each new one. Multi-Domain Learning (MDL) attempts to solve this problem by learning a single model capable of performing well in multiple domains. Nevertheless, the models are usually larger than the baseline for a single domain. This work tackles both of these problems: our objective is to prune models capable of handling multiple domains according to a user-defined budget, making them more computationally affordable while keeping a similar classification performance. We achieve this by encouraging all domains to use a similar subset of filters from the baseline model, up to the amount defined by the user's budget. Then, filters that are not used by any domain are pruned from the network. The proposed approach innovates by better adapting to resource-limited devices while being one of the few works that handles multiple domains at test time with fewer parameters and lower computational complexity than the baseline model for a single domain.
\end{abstract}

\begin{graphicalabstract}
\includegraphics[width=\textwidth]{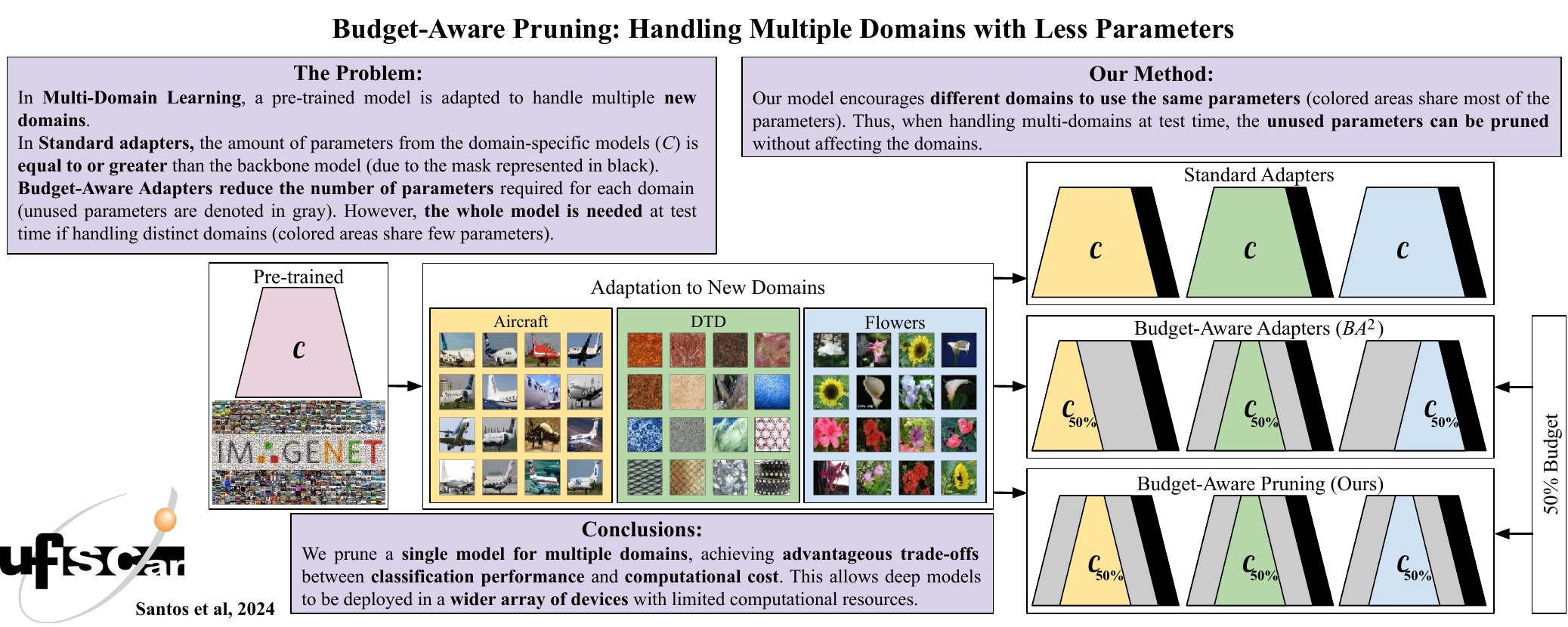}
\end{graphicalabstract}

\begin{highlights}
\item Multi-Domain Learning addressed with budget constraints for computational resources.
\item A method for sharing parameters among domains, allowing the pruning of unused ones.
\item Parameters are pruned according to a user-defined budget for computational resources.
\item A metric that considers the trade-off between accuracy, memory usage and computation.
\item Lower processing and memory costs than the original baseline for a single domain.
\end{highlights}


\end{frontmatter}

\section{Introduction}
\label{sec:introduction}

Deep learning has brought astonishing advances to computer vision, being used in several application domains, such as medical imaging~\cite{zhou2022review}, autonomous driving~\cite{wang2021rod2021}, road surveillance~\cite{nguyen2020anomaly}, and many others.
However, to increase the performance, 
increasingly deeper architectures have been used~\cite{liu2021group}, leading to models with a high computational cost. Also, for each new domain (or task) addressed, a new model is usually needed~\cite{berriel2019budget}. The significant amount of model parameters to be stored and the high GPU processing power required for using such models can prevent their deployment in computationally limited devices, like mobile phones and embedded devices~\cite{du2021bag}.
Therefore, specialized optimizations at both software and hardware levels are imperative for developing efficient and effective deep learning-based solutions~\cite{ISVLSI_2019_Marchisio}.

For this motive there has been a growing interest in learning multiple complex problems jointly, such as Multi-Task Learning (MTL)~\cite{aghajanzadeh2023task,sheng2021multi,chen2021residual,li2021multi,chen2022orthogonal,huang2022cross} and Multi-Domain Learning (MDL)~\cite{rebuffi2018efficient,mancini2020boosting,mallya2018piggyback,chattopadhyay2020learning,zhang2023alternative,wang2023mdl} methods.
These approaches are based on the observation that, although the tasks and domains can be very different, it is still possible that they share a significant amount of low and mid-level visual patterns~\cite{rebuffi2017learning}.
MTL extracts diverse information from a given sample, tackling multiple tasks, like object detection and segmentation.
While MDL aims to perform well in different visual domains, like medical images, images from autonomous vehicles, and handwritten text.
Both MTL and MDL goal is to learn a single compact model that performs well in several tasks/domains while sharing the majority of the parameters among them with only a few specific ones. This reduces the cost of having to store and learn a whole new model for each new task/domain.

Berriel~et~al.~\cite{berriel2019budget} point out that  one limitation of those methods is that, when handling multiple domains, their computational complexity is at best equal to the backbone model for a single domain. Therefore, they are not capable of adapting their amount of parameters to custom hardware constraints or user-defined budgets.
To address this issue, they proposed the modules named Budget-Aware Adapters (\textit{BA}$^2$) that were designed to be added to a pre-trained model to allow them to handle new domains and to limit the network complexity according to a user-defined budget.
They act as switches, selecting the convolutional channels that will be used in each domain.

However, as mentioned by Berriel~et~al.~\cite{berriel2019budget}, although this approach reduces the number of parameters required for each domain, the entire model still is required at test time if it aims to handle all the domains. The main reason is that they share few parameters among the domains, which forces loading all potentially needed parameters for all the domains of interest.

This work builds upon the \textit{BA}$^2$~\cite{berriel2019budget} by encouraging multiple domains to share convolutional filters, enabling us to prune weights not used by any of the domains at test time. Therefore, it is possible to create a single model with lower computational complexity and fewer parameters than the baseline model for a single domain. Such a model can better fit into a user's budget that has access to limited computational resources.

\begin{figure}[!htb]
    \includegraphics[width=\textwidth]{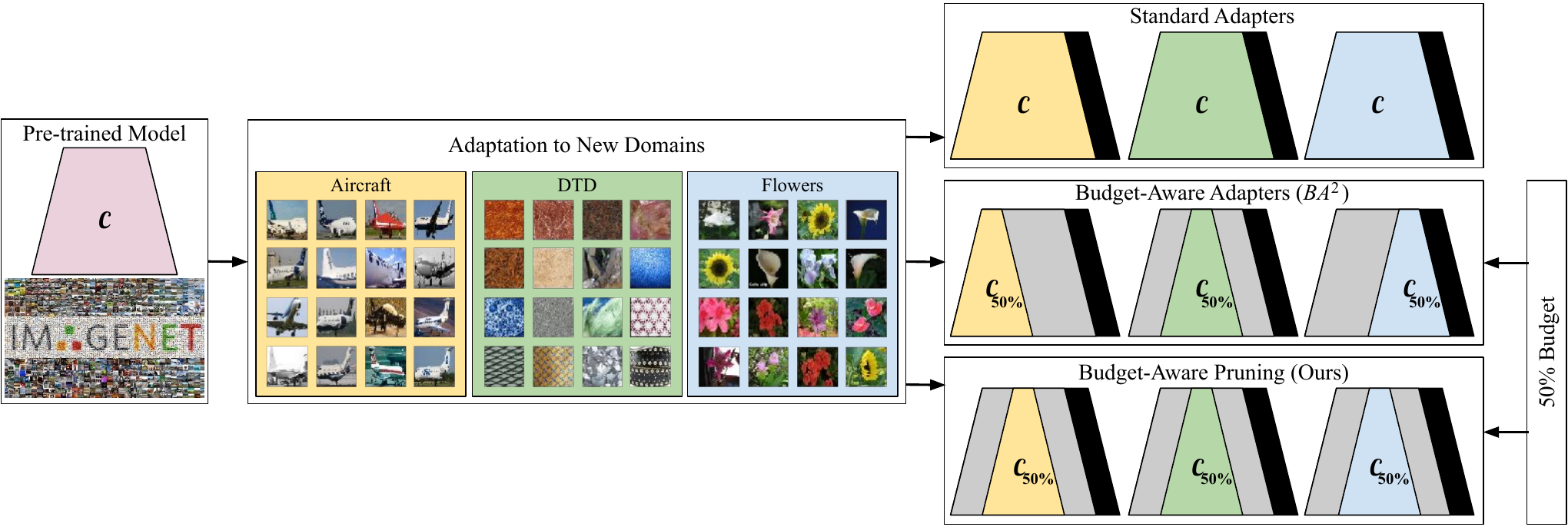}
    \caption{In MDL, a pre-trained model is adapted to handle multiple     domains. In standard adapters, the amount of parameters from the domain-specific models (indicated in colored $\mathcal{C}$) is equal to or greater than the backbone model (due to the mask represented in black).
    Budget-Aware Adapters~\cite{berriel2019budget} reduce the number of parameters required for each domain (unused parameters are denoted in gray). However, the whole model is needed at test time if handling distinct domains (colored areas share few parameters).
    Our model encourages different domains to use the same parameters (colored areas share most of the parameters). Thus, when handling multi-domains at test time, the unused parameters can be pruned.}
    \label{fig:visao_geral}
\end{figure}

Figure~\ref{fig:visao_geral} shows an overview of the problem addressed by our method, comparing it to previous MDL solutions and emphasizing their limitations.
As it can be seen, standard adapters use the entire model, while \textit{BA}$^2$~\cite{berriel2019budget} reduces the number of parameters used in each domain, but requires a different set of parameters per domain. Therefore, the entire model is needed for handling all the domains together and nothing can be effectively pruned.
On the other hand, our approach increases the probability of using a similar set of parameters for all the domains.
In this way, the parameters that are not used for any of the domains can be pruned at test time.
These compact models have a lower number of parameters and computational complexity than the original backbone model, which facilitates their use in resource-limited environments.
To enable the generation of compact models, we propose three novel loss functions that encourage the sharing of convolutional features among distinct domains.
Our 
approach was evaluated on two well-known benchmarks, the Visual Decathlon Challenge~\cite{rebuffi2017learning}, comprising 10 different image domains, and the ImageNet-to-Sketch setting, with 6 diverse image domains.
Results show that our proposed loss function is essential to encourage parameter sharing among domains, since without direct encouragement, the sharing of parameters tends to be low.
In addition, results 
show that our approach is comparable to the state-of-the-art methods in terms of classification accuracy, with the advantage of having considerably lower computational complexity and number of parameters than the backbone.

A preliminary version of this work~\cite{dos2023budget} was presented at the 22$^{nd}$ International Conference on Image Analysis and Processing (ICIAP 2023), where we introduced our intersection parameters-sharing loss function and manually selected the importance weight of the loss function.
In this work, we add several innovations.
We introduce two new Union and Jaccard parameter-sharing loss functions and use a strategy to enable learning the importance  weight for these losses as parameters of the model. These new strategies were crucial to increasing the performance of our method.
We introduce a novel metric, the efficiency-effectiveness score for multi-domain learning. With this metric, we can select the methods with the best trade-off between effectiveness on the classification task and the efficiency in computational complexity and number of parameters.
We also added more detailed experiments in the form of ablation studies.
Finally, we included additional comparisons with more state-of-the-art methods.

\section{Related Work}
\label{sec:related_work}


Previous MDL approaches used strategies like finetuning and pre-training, but faced the problem of catastrophic forgetting, in which the new domain is learned, but the old one is forgotten.
More recent approaches usually leverage a pre-trained model as the backbone. 
A considerable amount of the parameters from the backbone are shared for all the domains while attempting to learn a limited and much lower amount of new domain-specific parameters~\cite{berriel2019budget}.
Approaches mostly differ from each other according to the manner the domain-specific parameters are designed~\cite{berriel2019budget}.

Zhang~et~al.~\cite{zhang2023alternative} compared manually selecting the latter convolutional filters (top), the early filters (bottom), or random filters to be domain-specific. For MDL, making the early filters domain-specific obtained the best results by a considerable margin, which is different than most MTL approaches, where the top parameters are task-specific.

For methods that use residual blocks to learn new domains, an example is the work of Rebuffi, Bilen, and Vedaldi~\cite{rebuffi2017learning}, which adds domain-specific parameters to the ResNet network in the form of serial residual adapter modules, and another is the extended version presented in~\cite{rebuffi2018efficient} that proposes switching to parallel residual adapters. These modifications lead to an increase in accuracy and also a reduction in domain-specific parameters.
Iakovleva, Alahari, and Verbeek~\cite{iakovleva2023multi} propose to add modulation adapters to the convolutional layers, modifying their weights for each domain by a multiplicative relationship and reducing the number of new parameters added by decomposing them into two matrices with a smaller hidden dimension.

Following a different path, some works make use of binary masks to prune different convolutional filters of the network for each domain, like the Piggyback method proposed by Mallya, Davis, and Lazebnik~\cite{mallya2018piggyback}. 
The mask is initially learned with real values during training and then is thresholded to obtain binary values.
In test time, the learned binary mask is multiplied by the weights of the convolutional layer, keeping the value of some of them and setting the others to zero, generating a selection of different weights for each domain learned. Expanding on this idea, Mancini~et~al.~\cite{mancini2018adding} also make use of masks, however, unlike the method of Mallya, Davis, and Lazebnik~\cite{mallya2018piggyback}, which performs a multiplication, their approach learns an affine transformation of the weights through the use of the mask and some extra parameters. This work is further extended in~\cite{mancini2020boosting} by using a more general affine transformation and combining it with the strategy used by Mallya, Davis, and Lazebnik~\cite{mallya2018piggyback}. Focusing on increasing the accuracy with masks, Chattopadhyay, Balaji and Hoffman~\cite{chattopadhyay2020learning} propose a soft-overlap loss to encourage the masks to be domain-specific by minimizing the overlap between them. They were motivated by the fact that most domain generalization methods focus mainly on domain-invariant features, but domains usually have unique characteristics.

Due to the astonishing performance of transformers for computer vision tasks, some recent works also try to use them for MDL. Wang~et~al.~\cite{wang2023mdl} introduced MDL-NAS, a neural architectural method that searches for optimal architectures for each domain/task in two granularities. In the coarse granularity, the architecture of the model is searched, while in the fine granularity, parameters are shared among domains/tasks. Two parameter-sharing policies are used, sequential sharing, where parameters are shared for all tasks for each layer in order, and mask sharing, where different tasks can flexibly share parameters inside each layer.

The works mentioned so far mainly focused on improving accuracy while attempting to add a small number of new parameters to the model, but they do not take into consideration the computational cost and memory consumption, making their utilization on resource-limited devices difficult~\cite{yang2022da3}.
Trying to address that, recent works have attempted to tackle the MDL problem while taking into account resource constraints.

Senhaji~et~al.~\cite{senhaji2021not} proposed an adaptive parameterization approach that uses more parameters for more complex domains while using less for the easier ones.
To achieve this goal, parallel residual adapters~\cite{rebuffi2018efficient} are added to the ResNet-26, and then, the model is divided in three by adding early, mid and late exit points. Each domain, according to its complexity, can use the features of these exit points that are fed to a classifier.

Yang, Rakin, and Fan~\cite{yang2022da3} proposed the Dynamic Additive Attention Adaption (DA$^3$) to improve the training time on edge devices by reducing the activation memory used by the network. The authors achieve this goal by only updating the trainable parameters that have additive relationships with other weights or masks. Additive relationship calculations during backpropagation can be done independently from the activation feature maps from previous layers, which is not true for multiplicative relationships~\cite{yang2022da3}. Therefore, it is not necessary to store the activation of the entire network during training.

Regarding parameters sharing, Wallingford~et~al.~\cite{wallingford2022task} proposed the Task Adaptive Parameter Sharing (TAPS), which learns to share layers of the network for multiple tasks while having a sparsity hyperparameter $\lambda$ defined by the user.
This method learns what layers should be task-specific during the training of the model. If a layer is defined as task-specific, a residual is learned, that is a perturbation to be added to the weights of the layer. The sparsity hyperparameter $\lambda$ controls the amount of shared layers, where higher values encourage more layers to be shared amount tasks.
Although this method lessens the amount of additional domain-specific parameters, it still always has considerably more parameters than the backbone model for a single domain.

Berriel~et~al.~\cite{berriel2019budget} proposed Budget-Aware Adapters (BA$^2$), which are added to a backbone model, enabling it to learn new domains while limiting the computational complexity according to the user budget.
The BA$^2$ modules are similar to the approach from Mallya, Davis, and Lazebnik~\cite{mallya2018piggyback}, that is, masks are applied to the convolutional layers of the network, selecting a subset of filters to be used in each domain.
The masks are made of real values that are binarized during the forward pass but are used as real values during backpropagation.
The network is encouraged to use a smaller amount of filters per convolution layer than a user-defined budget, being implemented as a constraint to the loss function that is optimized by constructing a generalized Lagrange function. 
Also, the parameters from batch normalization layers are domain-specific, since they perform poorly when shared.

This method and other continual learning strategies can reduce the number of parameters for a single domain.
However, these methods usually load the relevant parameters for the desired domain at test time. In order to load them for each domain of interest, it would be necessary to keep all the parameters stored in the device so that the desired ones are available. This way, the model does not fit the user's needs, consuming more memory and taking more time to load, which might make it difficult to use in environments with limited computational resources. With this motivation, we propose our method that encourages the sharing of parameters.

\section{Pruning a Multi-Domain Model}
\label{sec:method}

\begin{figure}[!htb]
    \centering
    \includegraphics[width=0.8\textwidth]{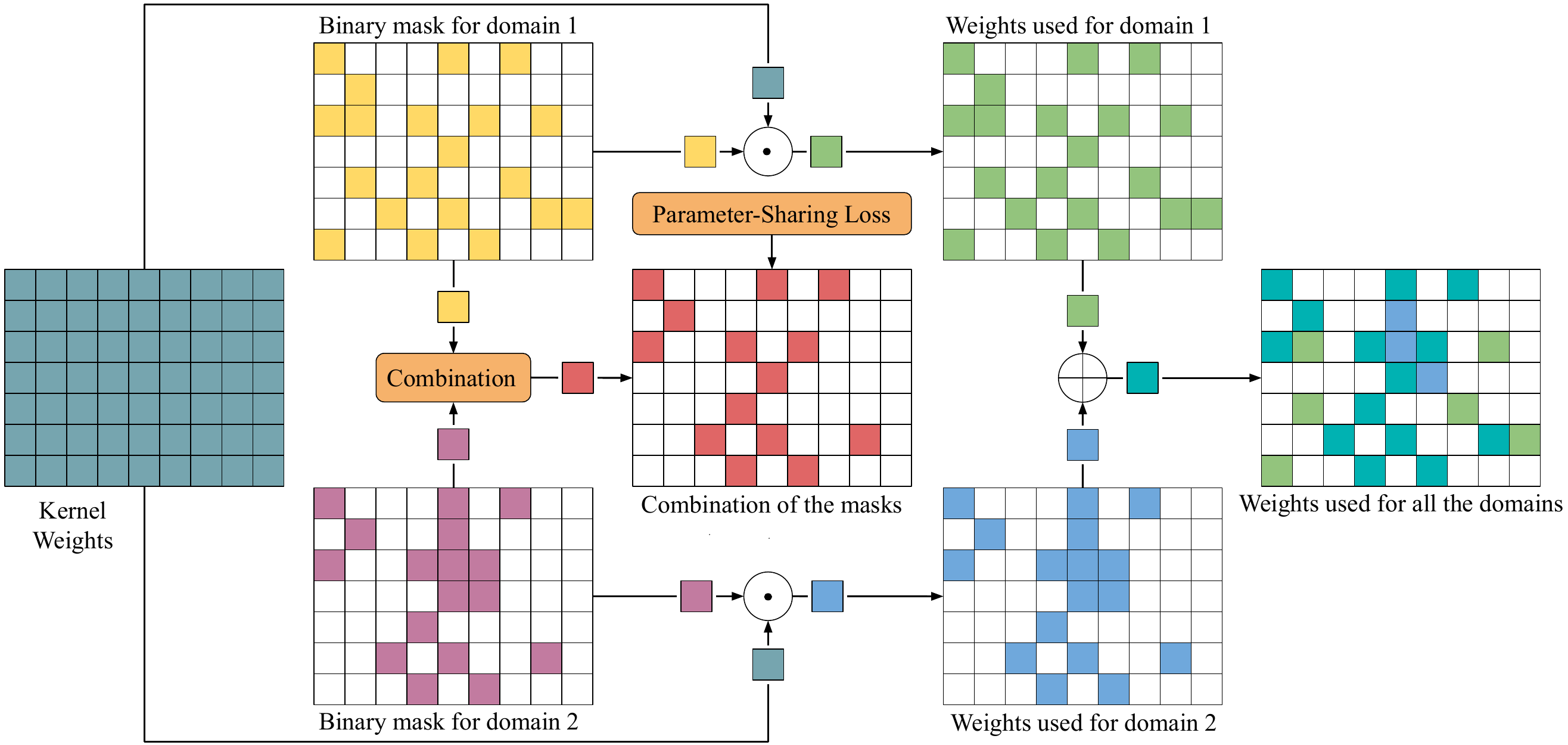}
    \caption{Overview of our strategy for sharing parameters among domains.
    Our parameter-sharing loss function is calculated over a combination of the masks from all the domains and is used to encourage the sharing of parameters between them.
    The parameters that are not used by any domain (white squares) can be pruned, reducing the number of parameters and computational cost of the model.
    $\odot$ represents the element-wise multiplication between a binary mask of a domain and the kernel weights and $\oplus$ represents the union of the weights used by each domain.
    Colors represent data (i.e., weights, masks, etc.), therefore, the colored squares denote both the input data for each operation as well as its resulting output.
    }
    \label{fig:method_overview}
\end{figure}

Our method was built upon the BA$^2$ modules from Berriel~et~al.~\cite{berriel2019budget} and proposes a new version to allow pruning unused weights at test time.
As results, the proposed method is able to obtain a pruned model that handles multiple domains, while having lower computational complexity and number of parameters than even the backbone model for a single domain. The pruned version is able to keep a similar classification performance while considering optimizations that are paramount for devices with limited resources.
Our user-defined budget enables the model to adapt to a wider range of environments based on available resources.
To achieve our goals, we added an extra loss function to BA$^2$ in order to encourage parameter sharing among domains and prune the weights that are not used by any domain. It was also necessary to train simultaneously in all the domains to be able to handle them all together at test time (see Figure~\ref{fig:method_overview} for an overview).

\subsection{Problem Formulation}
\label{sec:method_multi_domain}

The main goal of MDL is to learn a single model that can be used in different domains. One possible approach is to have a fixed pre-trained backbone model with frozen weights that are shared among all domains while learning only a few new domain-specific parameters.
Equation~\ref{eq:backbone} describes this approach, where $\Psi_0$ is the pre-trained backbone model that when given input data $x_0$ from the domain $X_0$ returns a class from domain $Y_0$ considering $\theta_0$ as the model's weights.
Our goal is to have a model $\Psi_d$ for each domain $d$ that attributes classes from the domain $Y_d$ to inputs $x_d$ from the domain $X_d$ while keeping the $\theta_0$ weights from the backbone model and learning as few domain-specific parameters $\theta_d$ as possible.
\begin{align}
    \label{eq:backbone}
    \Psi_0 (x_0; \theta_0): X_0 \rightarrow Y_0\\
    \Psi_d (x_d; \theta_0, \theta_d ): X_d \rightarrow Y_d\notag 
\end{align}

Our starting point was the BA$^2$~\cite{berriel2019budget} modules, which are associated with the convolutional layers of the network, enabling them to reduce their complexity according to a user-defined budget.
Equation~\ref{eq:conv} describes one channel of the output feature map $m$ at the location $(i,j)$ of a convolutional layer, where $g$ is the activation function, $K \in \mathbb{R} ^ {(2K_H + 1) \times (2K_W + 1) \times C}$ is the kernel weights with height of $2K_H + 1$, width of $2K_W + 1$ and $C$ input channels, and $I \in \mathbb{R} ^ {H \times W \times C} $ is the input feature map with $H$ height, $W$ width and $C$ channels.
\begin{align}
    \label{eq:conv}
    & m(i,j) = g( \sum^C_{c=1} \phi_c(i,j) )\\
    & \phi_c(i,j) = \sum^{K_h}_{h=-K_h} \sum^{K_w}_{w=-K_w} K(h, w, c) I(i-h, j-w, c)\notag 
\end{align}

Berriel~et~al.~\cite{berriel2019budget} proposed to add a domain-specific mask that is composed of $s_c$ switches for each input channel $c$, as shown in Equation~\ref{eq:ba2}.
At training time, $s_c \in \mathbb{R}$ while, at test time, they are thresholded to be binary values. When $s_c = 0$, the weights $K_c$ (i.e., the filters for the $c$ input channel for a given output channel) can be removed from the computational graph, effectively reducing the computational complexity of the convolutional layers.
\begin{align}
    \label{eq:ba2}
    m(i,j) = g( \sum^C_{c=1} s_c \phi_c(i,j) )
\end{align}

The model is trained by minimizing the total loss $L_{total}$, which is composed of the cross entropy loss $L$ and a budget loss $L_B$, as shown in Equation~\ref{eq:ba2loss}, where $\beta \in [0,1]$ is a user-defined budget hyperparameter that limits the number of weights on
each domain individually,  
$\theta_d^\beta$ are the domain-specific parameters for the budget $\beta$ and domain $d$, 
$\bar{\theta_d^\beta}$ is the mean value of the switches for all convolutional layers and  
$\lambda$ is the Karush-Kuhn-Tucker (KKT) multiplier.
In $L_B$, when the constraint $\bar{\theta_d^\beta} - \beta$ is respected, $\lambda=0$, otherwise, the optimizer increases the value of $\lambda$ to boost the impact of the budget.
\begin{align}
    \label{eq:ba2loss}
    L_B( \theta_d^\beta, \beta) = \max ( 0, \lambda ( \bar{\theta_d^\beta} - \beta ) )\nonumber\\
    L_{total} = L(\theta_0, \theta_d^\beta) + L_B( \theta_d^\beta, \beta)
\end{align}

\subsection{Sharing Parameters and Pruning Unused Ones}
\label{sec:method_sharing}

Although BA$^2$ can reduce the computational complexity of the model, it can not reduce the number of parameters necessary to handle all the domains together. Switches $s_c$ can only be pruned at test time when they are zero for \textit{all} domains, but they, in fact, assume different values if not forced to do so. 

For this reason, we added a parameter-sharing loss $L_{PS}$ to $L_{total}$, as we can see in Equation~\ref{eq:parameter_sharing}, where $N$ is the number of domains and $\theta_k^\beta$ for $k \in [1, ..., N]$ are the domain-specific parameters (switches or masks) for each domain.
\begin{align}
    \label{eq:parameter_sharing}
    L_{total} = L(\theta_0, \theta_d^\beta) + L_B( \theta_d^\beta, \beta) + L_{PS}( \theta_1^\beta, ..., \theta_N^\beta, \beta)
\end{align}

The parameter-sharing loss $L_{PS}$ calculates a combination of all the domains masks and uses it to encourage parameters to be shared among domains.
Since the domain-specific weights from all the domains are required by this loss component, it is necessary to train on all of them simultaneously. Finally, the switches $s_c$ and the associated kernel weights $K_c$ can be pruned.
It must be noted that the average sparsity over all domains may be higher than $1-\beta$, since
not all parameters are shared over all domains.

We proposed and tested three different parameter-sharing loss $L_{PS}$ functions:  $L_{PS}^{Int.}$, that is calculated using the intersection between the masks of different domains; $L_{PS}^{Union}$, that uses the masks union; and $L_{PS}^{Jaccard}$, that uses the Jaccard similarity coefficient.
In order to make these loss functions differentiable, the threshold operations are replaced by identity functions on backpropagation, following~\cite{mallya2018piggyback,berriel2019budget}.
Our model discards filters in the convolutional layers individually. When applied to models with depthwise separable convolutions, like the MobileNets~\cite{howard2017mobilenets}, where each convolution operation has one filter, the model can be pruned in a structured manner. When applied to models with standard convolutional layers, it is necessary to keep a table with the pruned weights in order to allow for the correct multiplication of weight matrices in a structured pruning scenario.

\subsubsection{Parameter-Sharing Intersection Loss}
\label{sec:method_sharing_intersection}

This was the first loss function we proposed~\cite{dos2023budget}, where we penalize the model when the intersection between the masks $\theta_k^\beta$ from all the domains $k \in [1, ..., N]$ have fewer switches than the maximum amount allowed by the budget, discouraging the use of different parameters by the domains.
The intersection was calculated by using element-wise multiplication between the binary masks from all the domains.
Equation~\ref{eq:parameter_sharing_intersection} describes this loss function, where $M$ is the total number of switches and $\lambda_{PS}^\beta$ is a weight that defines the importance of this loss component for the budget $\beta$.
\begin{align}
    \label{eq:parameter_sharing_intersection}
    L_{PS}^{Int.}( \theta_1^\beta, ..., \theta_N^\beta, \beta) = \max(0, \lambda_{PS}^\beta ( 1-  \frac{|\theta_1^\beta \cap \theta_2^\beta \cap ... \cap \theta_N^\beta|}{ M \beta}   ))
\end{align}

\subsubsection{Parameter-Sharing Union Loss }
\label{sec:method_sharing_union}

In this loss function, the model is penalized when the union between the masks $\theta_k^\beta$ from all the domains $k \in [1, ..., N]$ has more switches than the amount allowed by the budget $\beta$. 
This way, the loss function directly discourages the usage of more parameters than what the budget allows by all the domains jointly.  
Equation~\ref{eq:parameter_sharing_union} describes our parameter-sharing union loss $L_{PS}^{Union}$ function.
In order to calculate the union  between two binary masks $\theta_{i}^\beta$ and $\theta_{j}^\beta$ and make it differentiable, we followed Rahman~and~Wang~\cite{rahman2016optimizing}, where $\times$, $+$, and $-$ denote element-wise operations of multiplication, addition, and subtraction, respectively.
\begin{align}
    \label{eq:parameter_sharing_union}
    L_{PS}^{Union}( \theta_1^\beta, ..., \theta_N^\beta, \beta) = \max(0, \lambda_{PS}^\beta ( \frac{|\theta_1^\beta \cup \theta_2^\beta \cup ... \cup \theta_N^\beta|}{ M} - \beta   ))\\
    \theta_{i}^\beta \cup \theta_{j}^\beta = \theta_{i}^\beta + \theta_{j}^\beta - \theta_{i}^\beta \times \theta_{j}^\beta    
\end{align}

\subsubsection{Parameter-Sharing Jaccard Loss }
\label{sec:method_sharing_jaccard}

Chattopadhyay, Balaji, and Hoffman~\cite{chattopadhyay2020learning} proposed to use the Jaccard similarity coefficient~\cite{jaccard1901etude} in order to encourage masks to be domain-specific, sharing fewer parameters, being the opposite of our objective.
Motivated by this, we proposed a parameter-sharing Jaccard loss function that uses the complement of the Jaccard index in order to encourage the sharing of parameters.
This loss function does not take the budget $\beta$ as a parameter directly, this way, the budget loss $L_B$ needs to control the budget by itself, while our $L_{PS}^{Jaccard}$ only encourages the sharing of parameters.
\begin{align}
    \label{eq:parameter_sharing_jaccard}
    L_{PS}^{Jaccard}( \theta_1^\beta, ..., \theta_N^\beta) = \max(0, \lambda_{PS}^\beta ( 1-  \frac{|\theta_1^\beta \cap \theta_2^\beta \cap ... \cap \theta_N^\beta|}{|\theta_1^\beta \cup \theta_2^\beta \cup ... \cup \theta_N^\beta|}   ))
\end{align}

\section{Experiments and Results}
\label{sec:exp}

In this section, we present the experiments that were carried out and their results. First, we describe the experimental setup in detail. Then, we present ablation studies we made in order to train simultaneously in multiple domains and tune hyperparameters.
Finally, the results on the two well-known benchmarks, the Visual Decathlon Challenge and ImageNet-to-Sketch setting, are provided together with discussions.

\subsection{Experimental Setup}
\label{sec:exp:setup}

Our approach was validated on two well-known MDL benchmarks, the Visual Decathlon Challenge~\cite{rebuffi2017learning}, and the ImageNet-to-Sketch.
The Visual Decathlon Challenge comprises classification tasks on ten diverse well-known image datasets from different visual domains: 
ImageNet, Aircraft,  CIFAR-100, Daimler Pedestrian (DPed), Describable Textures (DTD), German Traffic Signs (GTSR), VGG-Flowers, Omniglot, SVHN, and UCF-101.
Such visual domains are very different from each other, ranging from people, objects, and plants to textural images.
The ImageNet-to-Sketch setting has been used in several prior works, being the union of six datasets: ImageNet, VGG-Flowers, Stanford
Cars, Caltech-UCSD Birds (CUBS), Sketches, and WikiArt~\cite{mallya2018piggyback}.
These domains are also very heterogeneous, having a wide range of different categories, from birds to cars, or art paintings to sketches~\cite{berriel2019budget}.

In order to evaluate the classification performance, we use the accuracy on each domain and the S-score~\cite{rebuffi2017learning} metric.
Proposed by Rebuffi Bilen and Vedaldi~\cite{rebuffi2017learning}, the S-score metric rewards methods that have good performance over all the domains compared to a baseline, and it is given by Equation~\ref{eq:s_score}:
\begin{equation}
    \centering
    \label{eq:s_score}
    S = \sum_{d=1}^{N} \alpha_d \max\{ 0, Err_{d}^{max} - Err_{d} \}^{\gamma_d} 
\end{equation}

\noindent where $Err_{d}$ is the classification error obtained on the dataset $d$,  
$Err_{d}^{max}$ is the maximum allowed error, that is the threshold beyond which points are no longer added to the score
and  $\gamma_d$ is a coefficient to ensure that the maximum possible $S$ score is $10.000$~\cite{rebuffi2017learning}.
In addition to the S-score, we also reported the classification accuracy on each domain and the mean accuracy over all of them.

To assess the computational cost of a model, we considered its amount of parameters and computational complexity.
For the number of parameters, we measured their memory usage, excluding the classifier and encoding float numbers in 32 bits and the mask switches in 1 bit.
For the computational complexity, we used the THOP library to calculate the amount of multiply-accumulate operations (MACs) for our approach\footnote{We follow Berriel~et~al.~\cite{berriel2019budget} and report results in FLOPs (1 MAC = 2 FLOPs).}.
For BA$^2$, we reported the values from \cite{berriel2019budget} and also contacted the authors to confirm that were using the same methods of measuring.

Similar to Berriel~et~al.~\cite{berriel2019budget}, in order to assess the trade-off between effectiveness on the MDL problem and computational efficiency, we consider two variations of the S-score, named S$_O$, which is the S-score per operation; and S$_P$, the S-score per parameter.
Also, we propose an efficiency-effectiveness score S$_E$ metric for MDL, allowing us to compare the overall performance of the models and select the best one among them.

Equations~\ref{eq:so}~and~\ref{eq:sp} describe the S$_O$ and S$_P$ metrics from Berriel~et~al.~\cite{berriel2019budget}. 
They indicate the ratio between the S-score and the computational complexity $O$ or the amount of parameters $P$. 
\begin{equation}
    S_O = \frac{S}{O}
    \label{eq:so}
\end{equation}
\begin{equation}
    S_P = \frac{S}{P}
    \label{eq:sp}
\end{equation}

In order to have a single metric taking into consideration both the computational complexity and number of parameters, we proposed our efficiency-effectiveness score S$_E$ metric, as we can see in Equation~\ref{eq:efficiency_score_mdl}, where both S$_O$ and S$_P$ are combined.
In this equation, S$^{baseline}_O$ and S$^{baseline}_P$ are the S$_O$ and S$_P$ values of the baseline model,  that is the method that uses a pre-trained model as a feature extractor, only training the classifiers.
\begin{equation}
    S_E = \frac{ S_O \times S_P  }{ S^{baseline}_O \times S^{baseline}_P }
    \label{eq:efficiency_score_mdl}
\end{equation}

For TAPS~\cite{wallingford2022task}, we used the implementation made available by the authors\footnote{TAPS implementation available at: \url{https://github.com/MattWallingford/TAPS}, as of June 2024.}. For the ImageNet-to-Sketch, we trained the available models and reported the results. For the Visual Domain Decathlon, we needed to adapt the code to run on the benchmark, since the original code did not make it available. In order to do so, we followed what is described on \cite{wallingford2022task} and contacted the authors for more details.
For this motive, there are some differences between the results reported in \cite{wallingford2022task} and what we obtained, but this process was necessary in order to guarantee that the measurements of the number of parameters and computational complexity used were the same.

In the ablation studies, we also reported the sparsity, representing the mean of the percentage per layer of filters that are not used by any of the domains and can be pruned.

For our methods, we adopted the same experimental protocol and hyperparameters as Berriel~et~al.~\cite{berriel2019budget}.
We used the SGD optimizer with a momentum of 0.9 for the classifier and the Adam optimizer for the masks. 
All weights from the backbone are kept frozen, only training the domain-specific parameters (i.e., classifiers, masks, and batch normalization layers) and the mask switches were initialized with the value of 10$^{-3}$.
Data augmentation with random crop and horizontal mirroring with a probability of 50\% was used in the training phase, except for DTD, GTSR,  Omniglot, and SVHN, where mirroring did not improve results or was harmful.
For testing, we used 1 crop for datasets with images already cropped (Stanford Cars and CUBS), five crops (center and 4 corners) for the datasets without mirroring, and 10 crops for the ones with mirroring (5 crops and their mirrors).
For the Visual Domain Decathlon, we used the Wide ResNet-28~\cite{zagoruyko2016wide} as the backbone, training it for 60 epochs with a batch size of 32, and learning rate of 10$^{-3}$ for the classifier and  10$^{-4}$ for the masks. Both learning rates are decreased by a factor of 10 on epoch 45.
For the ImageNet-to-Sketch setting, the ResNet-50 was used as the backbone, training for a total of 45 epochs with a batch size of 12, learning rate of 5$\times$10$^{-4}$ for the classifier and 5$\times$10$^{-5}$ for the masks, dividing the learning rates by 10 on epochs 15 and 30.

Differently from Berriel~et~al.~\cite{berriel2019budget}, we needed to train all the domains simultaneously, since we want to encourage the sharing of weights among them.
In order to do so, we run one epoch of each dataset in a round-robin fashion. We repeat this process until the desired number of epochs are reached for each dataset.

After obtaining the best hyperparameter configuration, the model was trained on both training and validation sets and evaluated on the test set of the Visual Domain Decathlon. 
Experiments were run using V100 and GTX 1080 TI NVIDIA GPUs, Ubuntu 20.04 distribution, CUDA 11.6, and PyTorch 1.12.

\subsection{Ablation Studies}
\label{sec:ablation}

In this section, we include ablation studies made while developing our method, providing more insights into the impact of some components of our proposal. A more detailed ablation studies and additional analysis are presented in the supplementary materials.

Initially, we tested running a version of our approach, in which all the domains are trained simultaneously, but without using the parameter-sharing loss (see supplementary materials). Compared to the original BA$^2$, with individual separate training for each domain, there was a small decrease in the mean accuracy after changing the training procedure to be simultaneous in all the domains, up to 2.4 percent.
This indicates that domains can affect each other during the training process and that small changes in the training procedures of the method can have effects on the results.
Nevertheless, it is necessary to keep this procedure in order to be able to use our loss function to share parameters among different domains, since it demands information from all the domains to work.
We also tested freezing the weights of all other domains except for the one from the input data, but our loss function was not able to encourage the sharing of parameters in this scenario.
Another aspect we tested was different strategies for performing the simultaneous training, like having one batch of each dataset, mixed batches with data from all domains, and a round-robin approach where we ran one epoch of each dataset in a random order.
We chose to use and report only the latter, as it was slightly faster and the accuracy of all were similar.

After evaluating the training of our method in all the domains simultaneously, we tested adding our different parameter-sharing loss functions to it.
In order to do so, we needed to define an appropriate value to the weight $\lambda_{PS}^\beta$, since it defines the impact of our parameter-sharing loss on the total loss.
We tested two strategies to achieve this objective.
Initially, we made a grid search with the fixed values of 1.0, 0.5, 0.25, and 0.125 and selected the best $\lambda_{PS}^\beta$ according to accuracy and sparsity for each $L_{PS}$ (see supplementary materials). 
Since some of our proposed loss functions did not achieve good results with this method, and manually testing a wider array of values would have a great computational cost, we tested setting $\lambda_{PS}^\beta$ as a parameter of the model, learning it jointly with the classification task.
In Table~\ref{tab:learned_lambda}, we compare the results of the best manually selected value of $\lambda_{PS}^\beta$ for each loss function with our approach of learning $\lambda_{PS}^\beta$ as a parameter of the model.

\begin{table}[!htb]
    \centering
    \caption{Comparison of mean accuracy and sparsity over all domains together for our method with the best manually selected value $\lambda_{PS}^\beta$ for each $L_{PS}$, and learning $\lambda_{PS}^\beta$ as parameters on the validation set of the Visual Decathlon Challenge.}
    \label{tab:learned_lambda}
    \footnotesize
    \begin{tabular}{lcccccc}
        \hline
        \hline
        \multirow{2}{*}{\textbf{Method}} & \multicolumn{2}{c}{\textbf{ $L_{PS}^{Int.}$~\cite{dos2023budget} }} & \multicolumn{2}{c}{\textbf{ $L_{PS}^{Union}$ }} & \multicolumn{2}{c}{\textbf{ $L_{PS}^{Jaccard}$ }}\\
        
         & \textbf{Acc.} & \textbf{Sparsity} & \textbf{Acc.} & \textbf{Sparsity} & \textbf{Acc.} & \textbf{Sparsity} \\
        \hline
        \hline

        \multicolumn{5}{l}{\textbf{Best with manually selected $\lambda_{PS}^\beta$:}}\\
        $\beta=1.00$ &  74.1 &  0.0 & \textbf{74.7} & 0.2 & 70.9 & 0.0 \\
        $\beta=0.75$ &  71.9 & 18.2 & 72.2 & 1.8 & 73.2 & 7.2 \\
        $\beta=0.50$ &  70.8 & 38.7 & 71.1 & 0.0 & 72.6 & 29.6 \\
        $\beta=0.25$ &  68.0 & 62.0 & 67.6 & 0.0 & 69.0 & 54.2 \\
        \hline

        \multicolumn{5}{l}{\textbf{Best with learned $\lambda_{PS}^\beta$:}}\\
        $\beta=1.00$ & 71.3 & 0.0  & 74.3 & 0.2  & 72.7 & 0.0 \\
        $\beta=0.75$ & 69.1 & 21.4 & 72.2 & 21.1 & 72.4 & 2.1 \\
        $\beta=0.50$ & 66.8 & 43.8 & 71.6 & 32.6 & 72.3 & 5.7 \\
        $\beta=0.25$ & 63.2 & \textbf{67.5} & 69.2 & 63.1 & 72.0 & 5.7 \\

        \hline
        
        \hline
    \end{tabular}
\end{table}

As we can see, for the best manually selected $\lambda_{PS}^\beta$ and most values of $\beta$, the Jaccard loss obtained the best accuracy, followed by the Union loss and the Intersection loss.
Although the Jaccard and Union losses obtained better accuracy than the Intersection loss, their sparsity over all the domains was lower.
The union loss obtained at best 1.8$\%$, achieving a very low sparsity for every configuration tested, this way, it would not be possible to prune the model and significantly reduce its cost.
The Jaccard loss function obtained better results, but they were still low when compared to the Intersection loss, which obtained the best trade-off between accuracy and sparsity.

When learning $\lambda_{PS}^\beta$, for the Jaccard loss, the sparsity was lowered when compared with the best manually selected $\lambda_{PS}^\beta$, obtaining overall low sparsity for all the budgets $\beta$.
While for the Intersection loss, there was an increase in sparsity with the learning values of $\lambda_{PS}^\beta$, but there was also a decrease in accuracy.
Finally, for the Union loss, the accuracy was similar between the best manually selected $\lambda_{PS}^\beta$ and the learned one, but the sparsity of the learned $\lambda_{PS}^\beta$ was high.
This way the Union loss function with learned $\lambda_{PS}^\beta$ obtained the best ratio between accuracy and sparsity and, for this reason, we selected it as our best configuration to be used in the next experiments presented in this work.

\subsection{Results on Visual Decathlon Challenge}
\label{sec:exp:results}

After selecting the best loss function and hyperparameter configuration, the model was trained on both training and validation sets and evaluated on the test set of the Visual Domain Decathlon. The comparison of our results with state-of-the-art methods is shown in Table~\ref{tab:acc_test}.
It must be noted that the sparsity hyperparameter $\lambda$ from TAPS decreases the number of parameters as its values increase, that is the inverse of our budget hyperparameter $\beta$, which decreases the number of parameters as its value is reduced. Also, as mentioned in Section~\ref{sec:exp:setup}, results have some differences from the ones reported in~\cite{wallingford2022task} because we needed to replicate the experiments on this benchmark since the implementation was not made available by the authors.

\begin{table*}[!htb]
	\centering
	\caption{Computational complexity, number of parameters, accuracy per domain, S, S$_O$, S$_P$, and S$_E$ scores on the Visual Domain Decathlon.}
	\label{tab:acc_test}
    \resizebox{\textwidth}{!}{
    \begin{tabular}{l|cc|cccccccccc|cccc}
        \hline
        \hline
        
        Method & FLOP & Params & ImNet & Airc. & C100  & DPed & DTD & GTSR & Flwr. & Oglt. & SVHN & UCF & S-score & S$_O$ & S$_P$ & {S$_E$}\\
        
        \hline
        \hline

        \multicolumn{12}{l}{{\textbf{Baselines~\cite{rebuffi2017learning}:}}}\\
        {Feature}  & {1.000} & {1.00}  & {59.7} & {23.3} & {63.1} & {80.3} & {45.4} & {68.2} & {73.7} & {58.8} & {43.5} & {26.8} &  {544} &  {544} & {544} & {1}\\
        {Finetune} & {1.000} & {10.0}  & {59.9} & \textbf{60.3} & {82.1} & {92.8} & {55.5} & {97.5} & {81.4} & {87.7} & {96.6} & {51.2} & {2500} & {2500} & {250} & {2.11}\\
        \hline

        \multicolumn{12}{l}{\textbf{TAPS~\cite{wallingford2022task}:}}\\
        $\lambda=0.25$ & 1.004 & 5.68 & \textbf{63.5} & 50.3 & \textbf{83.7} & 94.1 & 60.1 & 8.7 & 86.6 & 88.5 & 96.6 & 50.2 & 3113 & 3101 & 548 & {5.74}\\
        {$\lambda=0.50$} & {1.003} & {5.10} & \textbf{63.5} & {50.8} & {83.5} & {94.9} & {59.8} & {97.2} & {87.1} & {88.1} & {96.6} & {50.5} & {2821} & {2813} & {553} & {5.26}\\
        {$\lambda=0.75$} & {1.003} & {4.44} & \textbf{63.5} & {48.2} & {83.4} & {94.1} & \textbf{61.5} & {98.2} & \textbf{87.2} & {87.7} & {96.3} & \textbf{51.9} & {2919} & {2910} & {657} & {6.46}\\
        {$\lambda=1.00$} & {1.002} & {3.78} & \textbf{63.5} & {49.9} & {83.3} & {93.8} & {59.1} & {98.3} & {86.3} & {87.7} & {96.2} & {50.5} & {2862} & {2856} & {757} & {7.30}\\
        \hline

        \hline
        
        \multicolumn{12}{l}{\textbf{BA$^2$~\cite{berriel2019budget}:}}\\
        $\beta=1.00$ & 0.646 & 1.03  & 56.9 & 49.9 & 78.1 & 95.5 & 55.1 & 99.4 & 86.1 & 88.7          & \textbf{96.9} & 50.2 & \textbf{3199} & 4952 & 3106 & {51.97}\\
        $\beta=0.75$ & 0.612 & 1.03  & 56.9 & 47.0 & 78.4 & 95.3 & 55.0 & 99.2 & 85.6 & 88.8          & 96.8 & 48.7 & 3063 & 5005 & 2974 & {50.30}\\
        $\beta=0.50$ & 0.543 & 1.03  & 56.9 & 45.7 & 76.6 & 95.0 & 55.2 & 99.4 & 83.3 & \textbf{88.9} & \textbf{96.9} & 46.8 & 2999 & 5523 & 2912 & {54.35}\\
        $\beta=0.25$ & 0.325 & 1.03  & 56.9 & 42.2 & 71.0 & 93.4 & 52.4 & 99.1 & 82.0 & 88.5          & \textbf{96.9} & 43.9 & 2538 & 7809 & 2464 & {65.02}\\
        \hline

        \multicolumn{11}{l}{{\textbf{Ours:}}}\\

        {$\beta=1.00$} & {0.581} & {1.03}  & {56.9} & {48.7} & {77.9} & {95.5} & {55.1} & {99.2} & {85.1} & {88.5} & {96.7} & {47.6} & {3036} &  {5226} & {2948} & {52.06}\\
        {$\beta=0.75$} & {0.461} & {0.84}  & {56.9} & {43.4} & {76.8} & \textbf{95.7} & {54.0} & \textbf{99.5} & {78.3} & {88.0} & {96.6} & {44.3} & {2821} &  {6119} & {3358} & {69.43}\\
        {$\beta=0.50$} & {0.403} & {0.73}  & {56.9} & {43.1} & {76.2} & {95.4} & {52.1} & {99.0} & {76.7} & {88.2} & {96.4} & {44.1} & {2546} &  {6318} & {3488} & {74.47}\\
        {$\beta=0.25$} & \textbf{0.212} & \textbf{0.41}  & {56.9} & {35.2} & {69.8} & {95.4} & {48.5} & {98.8} & {71.4} & {87.3} & {96.3} & {41.9} & {2138} & \textbf{10085} & \textbf{5215} & {\textbf{177.72}}\\
        
        \hline
        \hline

    \end{tabular}
    }
\end{table*}

Compared to the baseline strategies, our method was able to vastly outperform the feature extractor only,
and compared to finetune, it was able to obtain a higher S-score for the budgets of $\beta=$1.0 and 0.75 and a similar S-score for $\beta=$0.50, but with almost 10 times fewer parameters.
The S$_E$ score from the baseline methods were also the lowest, since the feature extractor only had considerably low accuracy and S-score, while finetune is the method that uses the highest amount of parameters since an entire new model is used for each domain.
These results show the importance of the usage of MDL strategies in order to obtain a better balance between effectiveness and efficiency.

TAPS with $\lambda$ of 0.25 was the method with the second best S-score, losing only to BA$^2$ with $\beta=1.0$, but it also used more parameters than BA$^2$ and our method by a considerable margin, ranging from 3.78$\times$ to 5.68$\times$ times the size of the backbone model.
TAPS always uses considerably more parameters than the backbone model since it needs to learn the residuals that are added to the weights of some convolutional layers for each domain. Moreover, it also has a computational cost slightly higher than the backbone because of that.
For these reasons, TAPS only has higher S$_O$, S$_P$, and S$_E$ metrics than the baselines (i.e., feature and finetune), achieving values considerably lower than BA$^2$ and our strategy. 
This shows that the main focus of TAPS is different from our method and BA$^2$. The main objective of TAPS is to keep good accuracy, while adding a small amount of new parameters per new domain, while our work aims to prune the backbone model, reducing its computational complexity and number of parameters to fit a user-defined budget, with the smallest reduction in accuracy as possible.

BA$^2$ has a focus more similar to ours since it tries to reduce the computational complexity of the model according to the user budget, but it is not capable of pruning parameters. 
Since all domains together use almost all of them, it always has slightly more parameters than the backbone model. 
Compared to it, we obtained similar accuracy in most domains, facing drops in accuracy only for some domains.
We believe the main reason for this drop in accuracy is the simultaneous training procedure, as we observed a similar drop when switching from individual to simultaneous training without the addition of our loss function (as discussed in Section~\ref{sec:ablation}), but we kept it since it is necessary to enable parameter sharing.
The domains with the biggest accuracy drops were the smaller ones: aircraft, DTD, VGG-Flowers, and UCF-101.
Other works, like \cite{rebuffi2017learning,rebuffi2018efficient} also mention subpar performance on these datasets, identifying the problem of overfitting.

The S-score also dropped up to 453 points for the same issues. The drop is harsher since the metric was designed to reward good performance across all datasets, and the small datasets we mentioned had a subpar performance.
Despite facing small drops in accuracy and S-score, our method offers a good trade-off between classification performance and computational cost.

When comparing computational complexity (FLOP on Table~\ref{tab:acc_test}), our method obtained lower complexity than BA$^2$ for every budget. This happens due to the addition of our loss function which further encourages more weights to be discarded.
It also must be noted that all our methods obtained lower complexity than the value defined by the budget, showing that it is a great tool to adapt a backbone model to the resources available to users.

By comparing the S$_O$ metric, we can observe that both methods have a good trade-off between computational complexity and S-score, as this metric greatly increases as the budget is reduced, showing that the reduction in computational complexity is considerably greater than the loss in S-score. 
As expected, our method had better S$_O$ for every single budget $\beta$, since we also have a lower computational complexity.

The main advantage of our proposed method is the reduction in the number of parameters of the model, as it is, to our knowledge, it is one of the only methods that is capable of tackling the problem of multiple-domain learning, while also reducing the number of parameters in relation to the backbone model. Other methods can reduce the amount of parameters for a single domain, but since the parameters are not shared, to handle all of them during test time, the entire model must be kept.
As we can see (column Params of Table~\ref{tab:acc_test}), the original BA$^2$ had a similar amount of parameters to the backbone model, being 3\% more for all budgets.
For the budget of $\beta=$ 1.00, we obtained the same result, while for the budget of $\beta=$ 0.75, we reduced the amount of parameters compared to the backbone model by 16\%.
For the budget $\beta=$ 0.50, the reduction was 27\% and, for the budget of $\beta=$ 0.25, there were 59\% fewer parameters.

By comparing the S$_E$ score, we can also see the benefits of our models. Comparing with BA$^2$ with the same budgets, our method was better in all of them, being 0.09 points higher for $\beta=$ 1.00, 19.13 points higher for $\beta=$ 0.75, obtaining 20.12 more points for $\beta=$ 0.50 and for $\beta=$ 0.25 it was 112.7 points higher, being the highest S$_E$ score reported in this work.
Indeed, for all the budgets $\beta=$ 0.75 or less, we outperformed every configuration of every other method, showing that our strategy obtained the best trade-offs between efficiency and effectiveness.

These results show that our method was successful in encouraging the sharing of parameters among domains and that this approach can lead to considerable reductions in the amount of parameters of the network. 
The S$_P$ metric provides additional evidence to this finding, as for the budgets of $\beta=$ 0.75, 0.50, and 0.25, our method was able to outperform all configurations of BA$^2$ due to the considerable reduction in the amount of parameters.
The S$_E$ metric of our models was also the best among all compared works, indicating that our method was the one with the best balance between efficiency and effectiveness.

\subsection{Results on the ImageNet-to-Sketch Setting}

Table~\ref{tab:acc_benchmark2} compares our best method with state-of-the-art works on the ImageNet-to-Sketch setting.

\begin{table*}[!htb]
	\centering
	\caption{Computational complexity, number of parameters, accuracy per domain, S, S$_O$, S$_P$, and S$_E$ scores for the ImageNet-to-Sketch benchmark.}
	\label{tab:acc_benchmark2}
	\resizebox{\textwidth}{!}{
    \begin{tabular}{l|cc|cccccc|cccc}
        \hline
        \hline
        Method & FLOP & Params & ImNet & CUBS & Cars  & Flwr. & WikiArt & Sketches & S-score & S$_O$ & S$_P$ & {S$_E$} \\
        
        \hline
        \hline
        \multicolumn{8}{l}{{\textbf{Baselines:~\cite{mallya2018piggyback}:}}}\\ 
        {Feature}    & {1.000} & {1.00} & \textbf{76.2} & {70.7} & {52.8} & {86.0} & {55.6} & {50.9} &  {533} &  {533} & {533} & {1}\\
        {Finetune}  & {1.000} & {6.00} & \textbf{76.2} & \textbf{82.8} & {91.8} & \textbf{96.6} & {75.6} & \textbf{80.8} & \textbf{1500} & {1500} & {250}  & {1.32}\\
        \hline

        \multicolumn{8}{l}{\textbf{TAPS~\cite{wallingford2022task}:}}\\
        
        $\lambda=0.25$ & 1.046 & 4.32 & \textbf{76.2} & 82.2 & 89.3 & 95.9 & 80.1 & 79.2 & 1323 & 1265 & 306 & {1.36}\\
        $\lambda=0.50$ & 1.040 & 3.86 & \textbf{76.2} & 82.6 & 88.9 & 95.9 & \textbf{80.8} & 78.4 & 1315 & 1264 & 341 & {1.52}\\
        $\lambda=0.75$ & 1.032 & 3.31 & \textbf{76.2} & 82.6 & 88.3 & 95.3 & 79.8 & 79.1 & 1222 & 1184 & 369 & {1.54}\\
        $\lambda=1.00$ & 1.026 & 2.91 & \textbf{76.2} & 82.4 & 88.5 & 95.6 & 80.0 & 78.3 & 1240 & 1209 & 426 & {1.81}\\
        \hline

        \multicolumn{8}{l}{\textbf{BA$^2$~\cite{berriel2019budget}:}}\\
        $\beta=1.00$   & 0.700 & 1.03 & \textbf{76.2} & 81.2 & \textbf{92.1} & 95.7 & 72.3 & 79.3 & 1265 & 1807 & 1228 & {7.81}\\
        $\beta=0.75$   & 0.600 & 1.03 & \textbf{76.2} & 79.4 & 90.6 & 94.4 & 70.9 & 79.4 & 1006 & 1677 &  977 & {5.77}\\
        $\beta=0.50$   & 0.559 & 1.03 & \textbf{76.2} & 79.3 & 90.8 & 94.9 & 70.6 & 78.3 & 1012 & 1810 &  983 & {6.26}\\
        $\beta=0.25$   & 0.375 & 1.03 & \textbf{76.2} & 78.0 & 88.2 & 93.2 & 68.0 & 77.9 &  755 & 2013 &  733 & {5.19}\\
        \hline

        \multicolumn{8}{l}{{\textbf{Ours:}}}\\
        {$\beta=1.00$} & {0.612} & {1.07} & \textbf{76.2} & {80.8} & {91.9} & {92.4} & {75.0} & {79.5} & {1157} & {1890} & {1081}  & {7.19}\\
        {$\beta=0.75$} & {0.428} & {0.84} & \textbf{76.2} & {76.1} & {91.3} & {79.4} & {72.1} & {76.2} &  {889} & {2077} & {1058}  & {7.73}\\
        {$\beta=0.50$} & {0.590} & {0.59} & \textbf{76.2} & {73.8} & {90.7} & {83.8} & {69.6} & {75.1} &  {758} & {1284} & {1284}  & {5.80}\\
        {$\beta=0.25$} & \textbf{0.152} & \textbf{0.34} & \textbf{76.2} & {72.5} & {89.2} & {76.5} & {68.3} & {74.4} &  {641} & \textbf{4217} & \textbf{1885}  & {\textbf{27.98}}\\
        \hline
        
    \hline
    \end{tabular}}
\end{table*}

In the comparison with the baseline methods, as it can be seen, TAPS, BA$^2$, and our method were all able to vastly outperform the ``feature extractor only'' model in all metrics measured, while no method was able to obtain a better S-score than finetuning individual models for each domain, showing that ImageNet-to-Sketch is a challenging benchmark.

TAPS was the method with the best S-score and overall accuracy, but once again at the cost of using considerably more parameters than the backbone (ranging from 2.91$\times$ to 4.32$\times$ according to $\lambda$).
This way, their S$_O$, S$_P$, and S$_E$ metrics are again considerably worse than BA$^2$ and our method, reinforcing that, although TAPS also deals with the problem of taking into consideration the amount of parameters in multi-domain/task learning, its objective is very different from ours since it always needs to increase the amount of parameters in relation to the backbone.

Compared to BA$^2$, our method achieves a lower S-score for the same $\beta$, but in most cases, our model with the budget $\beta$ achieves a slightly higher S-score than the next lower $\beta$ from BA$^2$.
For example, our method with $\beta$=1.0 achieves an S-score 108 points lower than BA$^2$ with $\beta$=1.0, but it is 151 points higher than BA$^2$ with $\beta$=0.75.
For our method with $\beta$=0.50, the same pattern is repeated, achieving an S-score 254 points lower than BA$^2$ with $\beta$=0.50, but achieving 3 more points than BA$^2$ with $\beta$=0.25.
The only exception is for our method with $\beta$=0.75, where the S-score achieved was lower than BA$^2$ with $\beta$ of 0.75 and 0.50.
These results show that, although our method obtained slightly lower classification performance, it is still competitive, with the advantage of also being able to prune the model, reducing the number of parameters according to the user budget, something that $BA^2$ is not capable of doing.

With regard to the S$_O$ metric, our method obtained the best scores. For $\beta$=1.00, our method achieves higher S$_O$ than BA$^2$ with $\beta$ of 1.0, 0.75 and 0.25, and for the $\beta$ of 0.75, our method achieves S$_O$ score higher than every configuration of BA$^2$. This shows that our method is capable of obtaining a better trade-off between S-score and computational complexity than BA$^2$ in most cases.
Comparing the S$_P$ metric, since BA$^2$ can not prune the model, as $\beta$ decreases, S$_P$ is also reduced. In most cases for our methods, as the budget decreases,  S$_P$ increases, and for the $\beta$ of 0.50 and 0.25, we outperformed all the configurations of BA$^2$.
For the S$_E$ metric, our method was able to obtain better results for $\beta=$0.75 and 0.25, while for $\beta=$1.00 and 0.50, BA$^2$ was better. The overall best S$_E$ score was obtained by our method with a budget of $\beta=$0.75, since the computational complexity and amount of parameters were greatly reduced, but there was also a penalty on S-score.

The results obtained in this benchmark show that our method is capable of reducing both the number of parameters and computational complexity to fit a user budget while keeping competitive classification performance with other state-of-the-art methods. Our methods obtained the best trade-off between efficiency and effectiveness on the budgets of $\beta=$0.75 and 0.25, being the best approaches in these scenarios.

\section{Conclusions}
\label{sec:conclusions}

In this paper, we addressed the MDL problem while taking into account a user-defined budget for computational resources, a scenario addressed by few works, but of vital importance for devices with limited computational power.
One limitation of most of the MDL methods in literature is that, at best, their computational complexity is equal to the original single domain backbone~\cite{berriel2019budget}.
Since deep learning methods also tend to have high computational costs, this means that models for multiple domains would greatly benefit from strategies capable of reducing their complexity and amount of parameters, allowing for their deployment in a greater array of devices.

Berriel~et~al.~\cite{berriel2019budget} proposed the BA$^2$ modules, that select the channels that will be used in each domain respecting a user-defined budget restriction.
Although it is able to reduce the computational complexity, it is not capable of reducing the number of parameters, since different domains use different sets of parameters.
This limitation of the BA$^2$ method led us to our hypothesis that by encouraging parameter-sharing among domains, it would be possible to have a budget restriction capable of reducing both the computational complexity and amount of parameters.
We proved this hypothesis by proposing loss functions to encourage the sharing of parameters among domains, allowing us to prune the weights that are not used in any of them, reducing both the computational complexity and number of parameters to values lower than the original baseline for a single domain.

We proposed three different parameter-sharing loss functions based on different relationships between the masks that select the parameters of each domain.
The $L_{PS}^{Int.}$ loss takes in consideration the intersection between the masks, the $L_{PS}^{Union}$ used the union of the masks, and the $L_{PS}^{Jaccard}$ used the Jaccard similarity coefficient.
We also tested defining the importance of this loss manually and learning it as a parameter of the model.
The best overall results were obtained by $L_{PS}^{Union}$ learning its importance.

Performance-wise, our results were competitive with other state-of-the-art methods, while offering good trade-offs between classification performance and computational cost according to user needs.
In future works, we intend to evaluate different strategies for encouraging parameter sharing, apply our strategies to different network architectures, like the transformers~\cite{wang2023mdl}, and evaluate our methods on other benchmarks.

\section*{Acknowledgment}
{This work was supported by the FAPESP-Microsoft Research Institute (2017/25908-6), by the Brazilian National Council for Scientific and Technological Development - CNPq (310330/2020-3, 314868/2020-8), by LNCC via resources of the SDumont supercomputer of the IDeepS project, 
by the MUR PNRR project FAIR (PE00000013) funded by the NextGenerationEU and by EU H2020 project AI4Media (No. 951911).}

\bibliographystyle{IEEEtran}
\bibliography{bibliography.bib}

\begin{thebibliography}{10}
\providecommand{\url}[1]{#1}
\csname url@samestyle\endcsname
\providecommand{\newblock}{\relax}
\providecommand{\bibinfo}[2]{#2}
\providecommand{\BIBentrySTDinterwordspacing}{\spaceskip=0pt\relax}
\providecommand{\BIBentryALTinterwordstretchfactor}{4}
\providecommand{\BIBentryALTinterwordspacing}{\spaceskip=\fontdimen2\font plus
\BIBentryALTinterwordstretchfactor\fontdimen3\font minus \fontdimen4\font\relax}
\providecommand{\BIBforeignlanguage}[2]{{%
\expandafter\ifx\csname l@#1\endcsname\relax
\typeout{** WARNING: IEEEtran.bst: No hyphenation pattern has been}%
\typeout{** loaded for the language `#1'. Using the pattern for}%
\typeout{** the default language instead.}%
\else
\language=\csname l@#1\endcsname
\fi
#2}}
\providecommand{\BIBdecl}{\relax}
\BIBdecl

\bibitem{zhou2022review}
N.~Zhou, H.~Wen, Y.~Wang, Y.~Liu, and L.~Zhou, ``Review of deep learning models for spine segmentation,'' in \emph{Int. Conf. on Multimedia Retrieval (ICMR)}, 2022, pp. 498--507.

\bibitem{wang2021rod2021}
Y.~Wang, J.-N. Hwang, G.~Wang, H.~Liu, K.-J. Kim, H.-M. Hsu, J.~Cai, H.~Zhang, Z.~Jiang, and R.~Gu, ``Rod2021 challenge: A summary for radar object detection challenge for autonomous driving applications,'' in \emph{Int. Conf. on Multimedia Retrieval (ICMR)}, 2021, pp. 553--559.

\bibitem{nguyen2020anomaly}
K.-T. Nguyen, D.-T. Dinh, M.~N. Do, and M.-T. Tran, ``Anomaly detection in traffic surveillance videos with gan-based future frame prediction,'' in \emph{Int. Conf. on Multimedia Retrieval (ICMR)}, 2020, pp. 457--463.

\bibitem{liu2021group}
L.~Liu, S.~Zhang, Z.~Kuang, A.~Zhou, J.-H. Xue, X.~Wang, Y.~Chen, W.~Yang, Q.~Liao, and W.~Zhang, ``Group fisher pruning for practical network compression,'' in \emph{Int. Conf. on Machine Learning (ICML)}.\hskip 1em plus 0.5em minus 0.4em\relax PMLR, 2021, pp. 7021--7032.

\bibitem{berriel2019budget}
R.~Berriel, S.~Lathuillere, M.~Nabi, T.~Klein, T.~Oliveira-Santos, N.~Sebe, and E.~Ricci, ``Budget-aware adapters for multi-domain learning,'' in \emph{Int. Conf. on Comput. Vis. (ICCV)}, 2019, pp. 382--391.

\bibitem{du2021bag}
Y.~Du, Z.~Chen, C.~Jia, X.~Li, and Y.-G. Jiang, ``Bag of tricks for building an accurate and slim object detector for embedded applications,'' in \emph{Int. Conf. on Multimedia Retrieval (ICMR)}, 2021, pp. 519--525.

\bibitem{ISVLSI_2019_Marchisio}
A.~Marchisio, M.~A. Hanif, F.~Khalid, G.~Plastiras, C.~Kyrkou, T.~Theocharides, and M.~Shafique, ``Deep learning for edge computing: Current trends, cross-layer optimizations, and open research challenges,'' in \emph{{IEEE} Comput. Soc. Annu. Symp. on VLSI (ISVLS)}, 2019, pp. 553--559.

\bibitem{aghajanzadeh2023task}
E.~Aghajanzadeh, T.~Bahraini, A.~H. Mehrizi, and H.~S. Yazdi, ``Task weighting based on particle filter in deep multi-task learning with a view to uncertainty and performance,'' \emph{Pattern Recognition}, vol. 140, p. 109587, 2023.

\bibitem{sheng2021multi}
W.~Sheng and X.~Li, ``Multi-task learning for gait-based identity recognition and emotion recognition using attention enhanced temporal graph convolutional network,'' \emph{Pattern Recognition}, vol. 114, p. 107868, 2021.

\bibitem{chen2021residual}
B.~Chen, W.~Guan, P.~Li, N.~Ikeda, K.~Hirasawa, and H.~Lu, ``Residual multi-task learning for facial landmark localization and expression recognition,'' \emph{Pattern Recognition}, vol. 115, p. 107893, 2021.

\bibitem{li2021multi}
J.~Li, G.~Zhao, Y.~Tao, P.~Zhai, H.~Chen, H.~He, and T.~Cai, ``Multi-task contrastive learning for automatic ct and x-ray diagnosis of covid-19,'' \emph{Pattern recognition}, vol. 114, p. 107848, 2021.

\bibitem{chen2022orthogonal}
J.~Chen, L.~Yang, L.~Tan, and R.~Xu, ``Orthogonal channel attention-based multi-task learning for multi-view facial expression recognition,'' \emph{Pattern Recognition}, vol. 129, p. 108753, 2022.

\bibitem{huang2022cross}
N.~Huang, K.~Liu, Y.~Liu, Q.~Zhang, and J.~Han, ``Cross-modality person re-identification via multi-task learning,'' \emph{Pattern Recognition}, vol. 128, p. 108653, 2022.

\bibitem{rebuffi2018efficient}
S.-A. Rebuffi, H.~Bilen, and A.~Vedaldi, ``Efficient parametrization of multi-domain deep neural networks,'' in \emph{Conf. on Comput. Vis. Pattern Recognit. (CVPR)}, 2018, pp. 8119--8127.

\bibitem{mancini2020boosting}
M.~Mancini, E.~Ricci, B.~Caputo, and S.~Rota~Bul{\`o}, ``Boosting binary masks for multi-domain learning through affine transformations,'' \emph{Mach. Vis. and Applications}, vol.~31, no.~6, pp. 1--14, 2020.

\bibitem{mallya2018piggyback}
A.~Mallya, D.~Davis, and S.~Lazebnik, ``Piggyback: Adapting a single network to multiple tasks by learning to mask weights,'' in \emph{European Conf. on Comput. Vis. (ECCV)}, 2018, pp. 67--82.

\bibitem{chattopadhyay2020learning}
P.~Chattopadhyay, Y.~Balaji, and J.~Hoffman, ``Learning to balance specificity and invariance for in and out of domain generalization,'' in \emph{European Conf. on Comput. Vis. (ECCV)}.\hskip 1em plus 0.5em minus 0.4em\relax Springer, 2020, pp. 301--318.

\bibitem{zhang2023alternative}
L.~Zhang, Q.~Yang, X.~Liu, and H.~Guan, ``An alternative hard-parameter sharing paradigm for multi-domain learning,'' \emph{IEEE Access}, vol.~11, pp. 10\,440--10\,452, 2023.

\bibitem{wang2023mdl}
S.~Wang, T.~Xie, J.~Cheng, X.~Zhang, and H.~Liu, ``Mdl-nas: A joint multi-domain learning framework for vision transformer,'' in \emph{Conf. on Comput. Vis. Pattern Recognit. (CVPR)}, 2023, pp. 20\,094--20\,104.

\bibitem{rebuffi2017learning}
S.-A. Rebuffi, H.~Bilen, and A.~Vedaldi, ``Learning multiple visual domains with residual adapters,'' in \emph{Advances in Neural Information Processing Systems (NeurIPS)}, 2017.

\bibitem{dos2023budget}
S.~F. dos Santos, R.~Berriel, T.~Oliveira-Santos, N.~Sebe, and J.~Almeida, ``Budget-aware pruning for multi-domain learning,'' in \emph{Int. Conf. on Image Analysis and Processing (ICIAP)}, 2023, pp. 477--489.

\bibitem{iakovleva2023multi}
E.~Iakovleva, K.~Alahari, and J.~Verbeek, ``Multi-domain learning with modulation adapters,'' \emph{CoRR}, vol. abs/2307.08528, 2023.

\bibitem{mancini2018adding}
M.~Mancini, E.~Ricci, B.~Caputo, and S.~Rota~Bul{\`o}, ``Adding new tasks to a single network with weight transformations using binary masks,'' in \emph{European Conf. on Comput. Vis. (ECCV)}, 2018, pp. 0--0.

\bibitem{yang2022da3}
L.~Yang, A.~S. Rakin, and D.~Fan, ``Da3: Dynamic additive attention adaption for memory-efficient on-device multi-domain learning,'' in \emph{Conf. on Comput. Vis. Pattern Recognit. (CVPR)}, 2022, pp. 2619--2627.

\bibitem{senhaji2021not}
A.~Senhaji, J.~Raitoharju, M.~Gabbouj, and A.~Iosifidis, ``Not all domains are equally complex: Adaptive multi-domain learning,'' in \emph{Int. Conf. on Pattern Recognit. (ICPR)}.\hskip 1em plus 0.5em minus 0.4em\relax IEEE, 2021, pp. 8663--8670.

\bibitem{wallingford2022task}
M.~Wallingford, H.~Li, A.~Achille, A.~Ravichandran, C.~Fowlkes, R.~Bhotika, and S.~Soatto, ``Task adaptive parameter sharing for multi-task learning,'' in \emph{Conf. on Comput. Vis. Pattern Recognit. (CVPR)}, 2022, pp. 7561--7570.

\bibitem{howard2017mobilenets}
A.~G. Howard, M.~Zhu, B.~Chen, D.~Kalenichenko, W.~Wang, T.~Weyand, M.~Andreetto, and H.~Adam, ``Mobilenets: Efficient convolutional neural networks for mobile vision applications,'' \emph{CoRR}, vol. abs/1704.04861, 2017.

\bibitem{rahman2016optimizing}
M.~A. Rahman and Y.~Wang, ``Optimizing intersection-over-union in deep neural networks for image segmentation,'' in \emph{Int. Symp. on Visual Computing (ISVC)}.\hskip 1em plus 0.5em minus 0.4em\relax Springer, 2016, pp. 234--244.

\bibitem{jaccard1901etude}
P.~Jaccard, ``{\'E}tude comparative de la distribution florale dans une portion des alpes et des jura,'' \emph{Bull Soc Vaudoise Sci Nat}, vol.~37, pp. 547--579, 1901.

\bibitem{zagoruyko2016wide}
S.~Zagoruyko and N.~Komodakis, ``Wide residual networks,'' in \emph{British Mach. Vis. Conf. (BMVC)}, 2016.

\end{thebibliography}

\end{document}